\documentclass{article} 
\usepackage{iclr2026_conference,times}

\usepackage{amsmath,amsfonts,bm}

\def\eqref#1{equation~\ref{#1}}

\def\1{\bm{1}}

\DeclareMathAlphabet{\mathsfit}{\encodingdefault}{\sfdefault}{m}{sl}
\SetMathAlphabet{\mathsfit}{bold}{\encodingdefault}{\sfdefault}{bx}{n}

\usepackage{hyperref}
\usepackage{url}

\usepackage{graphicx}

\usepackage{booktabs}
\usepackage{tabularx}
\usepackage{siunitx} 
\usepackage{dblfloatfix} 

\usepackage{array}
\usepackage{makecell} 

\title{DexSIM: Real-time Dexterous Simulation with Unified Causal Video Diffusion}

\author{
Adam Lee \\
UC Berkeley \\
\texttt{alee00@berkeley.edu}
}

\iclrfinalcopy 

\begin{document}
\maketitle
\begin{abstract}
Recent progress of video diffusion models have enabled extensive simulation of the physical world. While simulation with hand object interaction has been less explored. We propose DexSIM, a dexterous simulation framework for simulating dexterous manipulation in real-time. While previous works utilizing video diffusion and 3D reconstruction focus on navigation, dexterous manipulation has been limited while it has extensive applications for creating interactive experiences with the simulated world and for generating synthetic data for robotics. Existing methods lack real-time interactivity and long-term spatial consistency and memory. We propose a 2-stage training framework for DexSIM. First we train a bi-directional video diffusion model by jointly embedding the hand action trajectory and video in a unified feature space. We utilize gaussian heatmap hand encoding for more accurate hand representation. Then we conduct a roll-out based autoregressive training with updated spatial cache as attention sink for spatial memory, which improves long-term consistency and 3D aware dexterous manipulation simulation. DexSIM outperforms the baseline on pixel and semantic similarity, motion fidelity, and hand projection accuracy. It also allows new applications such as hand motion transfer and runs at 15.24 FPS real-time interactivity.
\end{abstract}    
\section{Introduction}
Diffusion and flow models have allowed realistic generation of images and video~\cite{labs2025flux1kontextflowmatching, wan2025, videoworldsimulators2024}. With recent scaling and advancement of video diffusion has especially led to emergent abilities of simulating the physical world and interactions~\cite{videoworldsimulators2024}. Among generative simulation tasks dexterous manipulation simulation is an intuitive and natural extension of how we interact in the real world and thus allowing video diffusion models to simulate dexterous simulation can lead to natural interactions with the simulated world intuitively in creative applications. With long-term consistent and 3D aware simulations, dexterous generative simulations can also be used for training vision action models in robotics~\cite{huang2026pointworld}. While recent works have explored low latency real-time interactive causal video diffusion models for gaming~\cite{he2025matrix}, motion control~\cite{shin2025motionstream, zhao2025real}, and live avatar~\cite{huang2025live}, there hasn't been previous work for real-time interactive video generation for dexterous manipulation simulation.

Dexterous manipulation task aims to predict future states acting as a world model based on current state, given an action signal. As opposed to animation~\cite{wang2025unianimate}, hand inpainting tasks~\cite{chen2025foundhand}, the model must learn the dynamics its actions have on the environment and the more precise knowledge it has on the environment, input image, it achieves more physically realistic generations. This also differs from navigation tasks~\cite{he2025matrix, hunyuanworld2025tencent} which have much less changes in the environment and the dynamics and properties of interactions with different objects is not required. While previous work has focused on 2D signals~\cite{akkerman2025interdyn} and bi-directional approaches~\cite{kim2025dexterous}, we propose a new framework for causal generation with spatial cache to embody spatial awareness and memory. For manipulation, action conditions should be more precise, while 2D signals are limited to have ambiguity. While real time interactive has not yet explored for manipulation tasks which greatly hinders usage and adoption in commercial applications. We first train a bi-directional model that is most powerful. Then convert to an autoregressive model through self roll out with iterative training from teacher forcing to roll-out based Distribution Matching Distillation (DMD)~\cite{yin2024improved}  training. 

Prior works in dexterous manipulation~\cite{akkerman2025interdyn, kim2025dexterous, goswami2025world} face two significant challenges: lack of spatial awareness and consistency and real-time interaction. Spatial consistency is required for long-term consistency and accurate simulations since dexterous action conditions in 3D space can be ambiguous leading to inaccurate generation of hand object interactions. Also low-latency and real-time interactivity is required for creative applications such as augmented reality, whereas prior works lack causal generation or real-time interactive generation. 

To address these challenges we propose DexSIM, a 2-stage real-time causal generative simulation framework for dexterous manipulation. Given an input image DexSIM can generate video temporally autoregressively given streaming input of hand trajectories.

DexSIM is a real-time interactive causal video diffusion model with unified feature representation and spatial cache mechanism allowing real-time interactivity for dexterous manipulation simulation. First, we train a bi-directional video diffusion model that generates the entire frame sequence by jointly embedding the hand trajectory, initial 3D point and noise in the same space. Then we train our model with roll-out using DMD while including a updated spatial memory and initial scene attention sink to generate long-term consistent spatial aware interactive generation. Causal distillation results in a low-latency DexSIM with greater throughput with less number of function evaluations (NFE) resulting in a distilled 4-step model for real-time interactive dexterous manipulation simulation.  We rigorously evaluate DexSIM comparing to baseline on appearance similarity, and projected hand distance. 

Our contribution can be summarized as follows:
\begin{itemize}
  \item We propose DexSIM, a real-time interactive dexterous manipulation model based on a two-stage training framework with a unified feature space and gaussian heatmap hand encoding for more accurate hand representation.
  \item DexSIM generates real-time interactive simulations with spatial awareness and memory with an updatable spatial cache, resulting in a 15.24 fps interactive dexterous manipulation simulation.
  \item We rigorously evaluate DexSIM which outperforms the baseline on appearance quality, hand pose accuracy, and motion smoothness and allows new applications such as hand motion transfer.
\end{itemize}

\begin{figure}[t]
  \centering
  \includegraphics[width=\columnwidth]{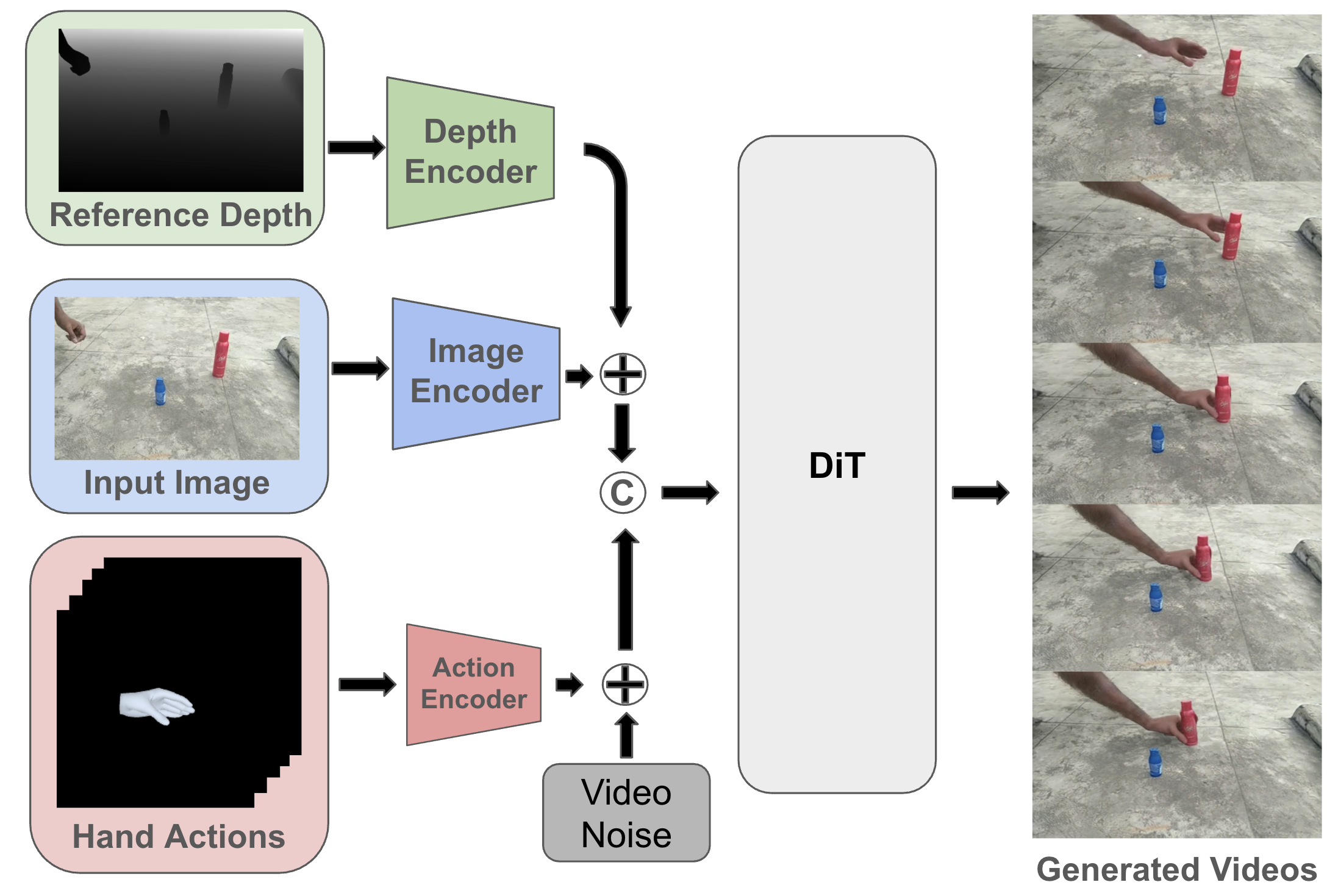}
  \caption{Overall architecture of DexSIM}
  \label{fig:arch}
\end{figure}
\section{Related Works}
\textbf{Causal Video Distillation} Flow matching is an ode solution to project from distribution p to q. Starting with image generation, video generation is enabled with an additional temporal dimension recent methods scaling DiT based architectures~\cite{videoworldsimulators2024, wan2025, kong2024hunyuanvideo}. Causal generation can be naively generated with teacher forcing but severly lacks long-term consistency. To address this challenge, diffusion forcing proposes progressively different levels of noise and history conditioning for video generation. Distribution Matching Distillation (DMD)~\cite{yin2024improved} initially proposed to utilize score matching from variational score distillation (VSD)~\cite{wang2023prolificdreamer} for step-distillation has been a crucial component for causal video distillation~\cite{yin2025slow}. Self-forcing~\cite{huang2025self} applies self-rollout to decrease train-test distribution gap. 

\textbf{Interactive Video Generation} LongLive conducts causal video generation with prompts but lacks specific detailed conditions. Drag simulation~\cite{zhao2025real, shin2025motionstream} has also been applied to synthesize motion from dragging but it lacks accurate 3D conditioning mostly relying on ambiguous 2D signals and lacks interactive capabilities with the 3D environment. Avatar lip syncing and simulation also focus on generating appearance wise realistic human pose and motion while interactions with surrounding objects and scene are limited. Interactive causal video generation has been applied to different applications in recent works. Including gaming primarily focused on scene navigation~\cite{he2025matrix, kong2024hunyuanvideo}. While these models are impressive and can conduct 3D tasks such as novel view synthesis one-shot, it lacks the ability to physically interact with the environment where the dynamics and forces are applied to different surrounding objects. 


\textbf{Dexterous Manipulation}
InterDyn~\cite{akkerman2025interdyn} generates video with bi-directionally based on 2D segment conditioning trained on controlnet. DWM~\cite{kim2025dexterous} generates video bi-directionally trains a video diffusion model to learn from hand warped video. DexWM~\cite{goswami2025world} is trained from hand pose difference as action signals to a causal video generation model. It has 3D signal for hand action but lacks 3D awareness of scene and memory. Also lacks real-time interactivity and long-term consistency due to train test time gap addressed by DMD and self-forcing. EgoEdit~\cite{li2025egoedit} performs egocentric hand streaming input video can interact with objects based on text input but lacks ability to interact with given 3D, generates hand interacting augmented reality. Since in robotics tasks and for 3D scene interaction, accurate 3D action conditioning and spatial awareness is required for the model to learn which objects are simulated it's dynamics magnitude. While other works have proposed non-egocentric simulation words, PlayerOne~\cite{tu2025playerone} the realistic interaction is not as much of a priority as egocentric simulation and serves as a novel view synthesis task. EgoEdit generates egocentric simulation based on text context which also has limited explicit action conditioning and the focus is less on accurate interaction with the given environment but generating synthetic objects and assets for interaction that matches the text prompt. Interactivity requires better spatial consistency and awareness. Since it interacts with the 3D world and objects from the given input. Hand inpainting~\cite{chen2025foundhand} and animation generation~\cite{wang2025unianimate} are similar in 2D key point conditioning, these tasks focus on accurate pose generation without learning to simulate the dynamics between the pose and the environment.
\section{Preliminaries}
\subsection{Video Diffusion}
  Diffusion models learn a reverse-time denoising process that transforms a simple noise distribution into the data distribution.
Let $\mathbf{x}_0 \sim p_d$ denote a data sample. A forward noising process $q$ length $T$ has a variance schedule $\{\beta_t\}_{t=1}^T$, where $\alpha_t = 1-\beta_t$ and $\bar{\alpha}_t = \prod_{s=1}^t \alpha_s$.
The forward process admits the closed form
\begin{equation}
q(\mathbf{x}_t \mid \mathbf{x}_0)
= \mathcal{N}\!\left(\mathbf{x}_t;\ \sqrt{\bar{\alpha}_t}\,\mathbf{x}_0,\ (1-\bar{\alpha}_t)\mathbf{I}\right),
\end{equation}
equivalently,
\begin{equation}
\mathbf{x}_t = \sqrt{\bar{\alpha}_t}\,\mathbf{x}_0 + \sqrt{1-\bar{\alpha}_t}\,\boldsymbol{\epsilon},
\qquad \boldsymbol{\epsilon}\sim\mathcal{N}(\mathbf{0},\mathbf{I}),\ \ t\in\{1,\dots,T\}.
\label{eq:forward_diffusion}
\end{equation}

A neural network ${\epsilon}_\theta$ is trained to predict a denoising target from a noisy sample $\mathbf{x}_t$ and timestep $t$, and a conditioning $c$.
Using the common noise-prediction parameterization, the objective is
\begin{equation}
\mathcal{L}_{\text{DDPM}}
=
\mathbb{E}_{t\sim p(t),\ \mathbf{x}_0\sim p_d,\ \boldsymbol{\epsilon}\sim \mathcal{N}(\mathbf{0},\mathbf{I})}
\Big[
w(t)\, \big\|\boldsymbol{\epsilon}_\theta(\mathbf{x}_t, t, c) - \boldsymbol{\epsilon}\big\|_2^2
\Big],
\label{eq:ddpm_loss}
\end{equation}

Modern implementations conduct the diffusion process in the latent space with a diffusion transformer (DiT)~\cite{peebles2023scalable} backbone. And the denoised latent is decoded to pixels through a 3D VAE. 

\subsection{Distribution Matching Distillation}
  Distribution Matching Distillation (DMD) utilizes the score difference between teacher and fake model given the noised latent of a fully denoised latent from the generator. 
  \begin{equation}
\nabla \mathcal{L}_{\mathrm{DMD}}
= -\mathbb{E}_t\!\left(
\int \Big(
s_{\mathrm{real}}\!\big(F(G_\theta(z),t),t\big)
-
s_{\mathrm{fake}}\!\big(F(G_\theta(z),t),t\big)
\Big)\,
\frac{dG_\theta(z)}{d\theta}\,dz
\right).
\end{equation}

  The generator can be a causal model while the teacher is bi-directional, more expressive bigger model and the fake model updates more frequently to match generator distribution. As a distribution matching objective VSD was originally proposed for 3D distillation but DMD is utilized for step distillation while recent works~\cite{yin2025slow, huang2025self} have adapted it for causal video distillation. Other distillation and joint distillation of different properties such as cfg, step, and causal video is possible and remains a future work to inspect the optimal settings for joint distillation. 

\begin{figure*}[t]
  \centering
  \includegraphics[width=1.0\textwidth]{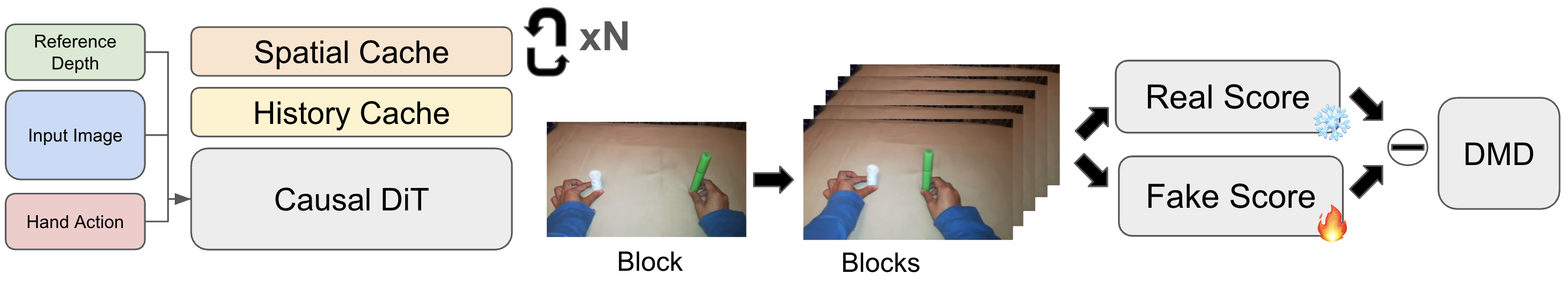}
  \caption{Rollout training with Spatial Memory}
  \label{fig:rollout}
\end{figure*}

\section{Method}
DexSIM predicts the future frames given an input image and hand action conditions, and depth as a world model. It has to learn to model the hand and naturally interact with the objects in the given input image and environment using pre-trained priors and depth information. We focus on egocentric views to model how people interact with objects in the real world. In the following sections we describe our architecture and the 2-stage training process with spatial cache.

\subsection{Task formulation}
Given a state $S$, we aim to predict the future states $\{s_t\}_{t=1}^T$ conditioned on actions $\{c_t\}_{t=1}^T$ with $c_t \in \mathcal{C}$. We represent each state by a latent image encoding $s_t := \mathrm{enc}(I_t)$. For the first-stage bi-directional training stage, DexSIM predicts all future states immediately:
\begin{equation}
p\!\left(s_{1:T}\mid s_0,\, c_{1:T},\, d_0\right),
\end{equation}

where $d_0$ is the depth latent of the initial state. And DexSIM predicts all future states $s_{1:T}$ at once. Causal DexSIM from the second stage training aims to predict the next state autoregressively resulting in the following formulation:
\begin{equation}
p\!\left(s_t \mid s_{<t},\, c_{\le t},\, d\right).
\end{equation}

Here $c_t$ is implemented as sliding window attention with $n$ most recent frames and first frame initial sink. $d$ represents the updatable spatial cache which is described in the following section.

\subsection{Architecture}
We employ a bi-directional video diffusion model as our base model and train by jointly embedding the hand trajectory and initial 3D map to condition the generation.

We adopt a hand action representation framework inspired by the hand inpainting domain. As mentioned in~\cite{chen2025foundhand} 3D hand representations such as Mano are more expensive to gather and also less robust to occlusion. To provide spatial signals we include the initial depth map in the image encoder. 

\textbf{Unified feature representation} Inspired by animation literature~\cite{wang2025unianimate} we adopt a unified feature representation 
As hand action signal is crucial and requires long term consistency. We adopt a DiT architecture with a unified feature space with the condition and noise. While previous work~\cite{akkerman2025interdyn} and Controlnet~\cite{zhang2023adding} employ a dual model for conditioning or dual network for joint prediction facilities feature alignment with less parameters. Memory optimization is crucial for 2nd stage causal distillation and DMD. 

While previous methods implicitly learn relevant state of the environment including current pose and the spatial distance, we incorporate the references with separate encoders. The target action is encoded and its feature is fused with the noise latent. 

\textbf{Hand action and state conditioning} We use mediapipe landmark keypoints for both hands resulting in keypoints $p \in \mathbb{R}^{42 \times 2}$ and project it to gaussian heatmap following~\cite{chen2025foundhand}:
\begin{equation}
\mathcal{H} = \mathcal{N}(x \mid p, \sigma^2),
\end{equation}
resulting in 42 channel hand representation  $\mathbf{h}_t \in \mathbb{R}^{42 \times h \times w}$ which we encode with a convolutional encoder resulting in a hand action latent $\mathbf{l}_t \in \mathbb{R}^{c \times h' \times w'}$. The size of latent channel $c$ is set to 1536. With hand positioning being a crucial signal we follow ~\cite{wang2025unianimate} and add the hand pose to the initial noise. This hand representation is much more expressive in representing hand pose than previous work ~\cite{akkerman2025interdyn} relying on 2D masks. We adopt a depth map of the initial frame and project it to the joint space as the latent space. This state acts as a guidance to infer spatial dynamics and action correspondences complementing the 2D hand representations. 

To increase accurate hand pose projection and generation we refer to the hand inpainting literature~\cite{chen2025foundhand} and employ a temporal hand trajectory projection, projection 22 keypoints from both hands into noise token space. 

Our model is trained on the latent space $\mathbf{z}_0 \in \mathbb{R}^{f \times c \times h \times w}$ with standard diffusion loss with depth $c_h$ and hand $c_h$ conditions:
\begin{equation}
\mathcal{L}_{\mathrm{LDM}}
= \mathbb{E}_{z_0,t,\epsilon}\!\left[ \left\| \epsilon - \epsilon_{\theta}\!\left(z_t, t, \mathbf{c}_d, \mathbf{c}_h\right) \right\|_2^2 \right].
\end{equation}
During inference, hand from source streaming video is projected to source image and an initial 3D map is utilized for spatial awareness while simulating from hand trajectories that require inductive bias to accurately simulate surrounding object manipulation.

\begin{figure}[t]
  \centering
  \includegraphics[width=\columnwidth]{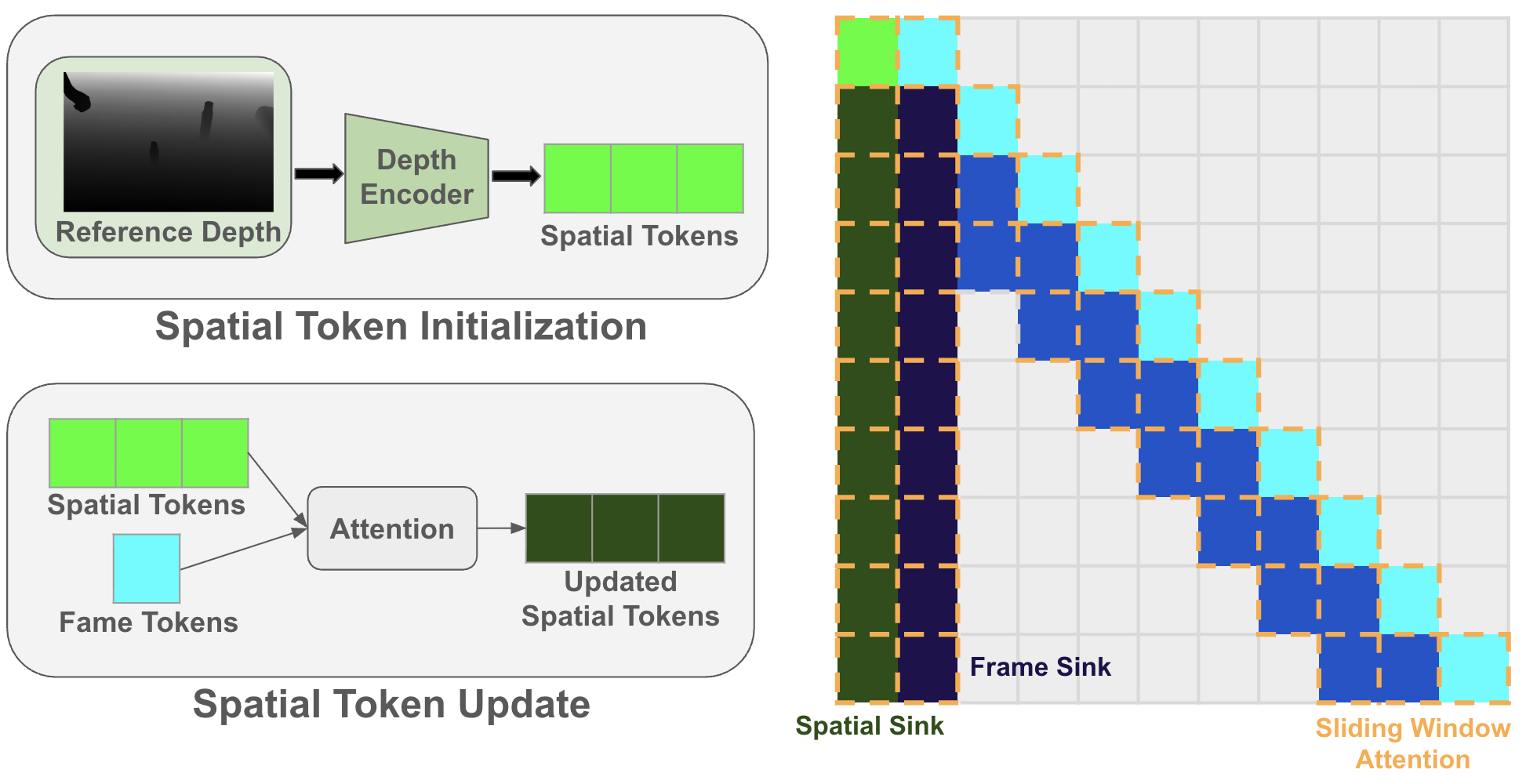}
  \caption{Spatial Cache}
  \label{fig:sm}
\end{figure}

\subsection{Roll-out training with Spatial cache}
The main challenges with autoregressive generation is long term consistency which is more challenging for dexterous interaction generation as they require spatial understanding to generate realistic simulations and accurate interactions with the environment. Whereas, for dexterous manipulation simulation, real-time interactivity has been the bottleneck for usage in creative applications and robotics tasks requiring real-time guidance. To address these challenges we first train the model with teacher forcing with causal attention. Then train with self roll out based DMD loss where the bi-directional DexSIM is used as the teacher model. 

We adopt a spatial cache that acts as an attention sink along with the frame sink used in causal video generation~\cite{yang2025longlive, huang2025live}. The spatial cache is updated with new blocks with cross-attention. We encode the initial depth of the scene to project to generate key and value tokens. The tokens are subsequently updated after a new block is generated from the causal DiT with cross attention. This updated spatial guidance allows the model to have long term spatial awareness and memory.

We design an updatable spatial cache which is updated with frame-wise autoregressive video generation during inference. Initialized by the 3D map used as the training condition for the bi-directional model $d \in \mathbb{R}^{1 \times h \times w}$, we project the depth map using a convolutional encoder to match the embedding space of the DiT latents resulting in a depth latent $d' \in \mathbb{R}^{n \times c}$ where $n$ is the number of tokens and $c$ is the channel. In practice, we use 256 tokens and 1536 channels. They are updated by the initial frame latent tokens with cross-attention and for each subsequently generated frame latent blocks, we update the spatial cache tokens for more accurate 3D consistent generation. This update allows the depth conditioning to modify based on the new frames and it's changes from the previous frames. 

We adopt the initial 3D map as condition through attention sink along with updatable spatial memory to preserve long-term 3D consistency and awareness along with providing with an initial spatial reference point. Resulting in a total 512 additional tokens that are used as conditions for causal frame generation. Through both updatable spatial memory and initial scene attention sink, DexSIM is capable of depth aware dexterous manipulation. Autoregressive generation allows interactive and real time interactions with the conditioned environment. 

\begin{table*}[!t]
\centering
\small
\setlength{\tabcolsep}{4pt}
\renewcommand{\arraystretch}{1.1}

\begin{tabularx}{\textwidth}{l *{5}{>{\centering\arraybackslash}X}}
\toprule
Method & SSIM $\uparrow$ & PSNR $\uparrow$ & LPIPS $\downarrow$ &
Motion Fidelity $\uparrow$ & PCK@20 $\uparrow$ \\
\midrule
InterDyn & 0.506 & 11.64 & 0.388 & 0.576 & 61 \\
DexSIM Bi-directional & 0.573 & 15.92 & 0.322 & 0.613 & 71 \\
DexSIM Causal & 0.526 & 12.21 & 0.362 & 0.594 & 65 \\
\bottomrule
\end{tabularx}
\caption{Comparison with baseline . $\uparrow$ higher is better, $\downarrow$ lower is better.}
\label{tab:main}
\end{table*}

\begin{figure*}[t]
  \centering
  \includegraphics[width=1.0\textwidth]{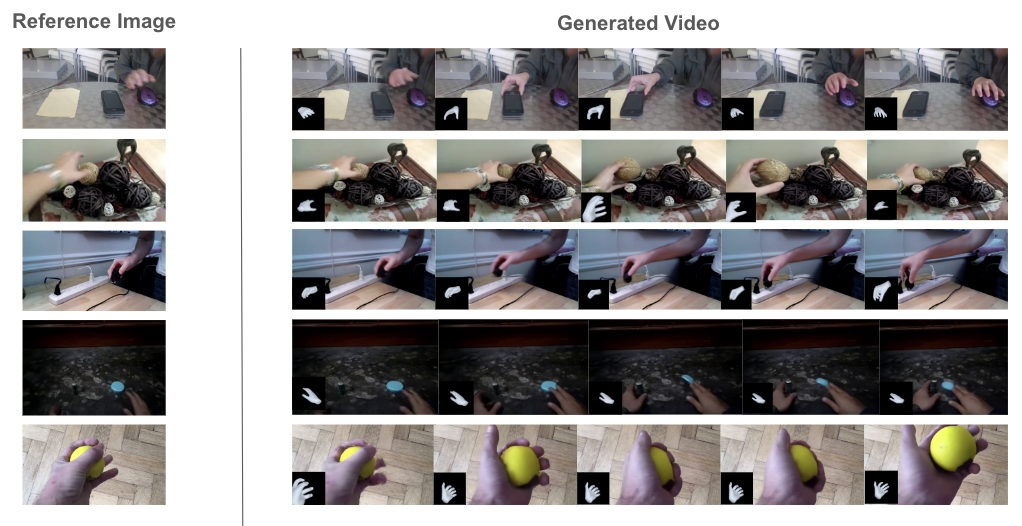}
  \caption{Qualitative Results}
  \label{fig:mc_1}
\end{figure*}

\section{Experiments}

\subsection{Experimental settings}
We initialize our model with a Wan 1.3B base model~\cite{wan2025}. The the action encoders are implemented with convolutional layers. The experiments are done in 8 H100 GPUs. Number of frames is set as 81 with spatial resolution resized to 768 x 512 with 5k steps for stage 1 and 2k steps for stage 2. Per GPU batch size is 4 for stage 1 and 1 for stage 2. We follow self-forcing to reduce memory footprint by using adopting gradient accumulation and stop gradient for each step. We train both stages with LoRA~\cite{hulora} with rank 64. 

We employ hand object interaction dataset something-something-v2 (ssv2) to train our model. To filter for quality we filter out frames that do not have hand action signals and to mitigate only sample video chunks with a continuous hand signal where a hand signal not present for 12 consecutive frames is discarded and a new chunk is initialized when they hand is visible again. 

\textbf{Metrics}
While appearance and motion smoothness rely on scaling model, data and compute, DexSIM focuses on hand pose accuracy and motion fidelity of generated simulations. For evaluation we use the test set of the ssv2 dataset which contains hand object interaction videos at scale the majority at a static setting. 

We compare with the baseline InterDyn which similarly generates videos of forced conditioned based on a 2D mask. We acquire the segmentation by getting the keypoints and using them as a prompt for SAM to generate the condition masks. We utilize both perceptual and pixel level similarity metrics as well as PCK@20 which evaluates the amount of keypoints that are within 20 pixels of target. We use a Cotracker~\cite{karaev24cotracker3} based motion fidelity metric by ~\cite{yatim2024space} to measure motion fidelity. We follow a similar implementation to ~\cite{akkerman2025interdyn} and track the pixels of the hand masks from the initial image with two tracklets from the ground truth and the comparison model \(T=\{\tau_1,\ldots,\tau_n\}\) and \(\tilde{T}=\{\tilde{\tau}_1,\ldots,\tilde{\tau}_m\}\).. 
\begin{equation}
\frac{1}{m}\sum_{\tilde{\tau}\in \tilde{T}}\max_{\tau\in T}\mathbf{corr}(\tau,\tilde{\tau})
\;+\;
\frac{1}{n}\sum_{\tau\in T}\max_{\tilde{\tau}\in \tilde{T}}\mathbf{corr}(\tau,\tilde{\tau}),
\end{equation}
where the correlation $corr$ is defined as the following:
\begin{equation}
\mathbf{corr}(\tau,\tilde{\tau})
=
\frac{1}{F}\sum_{k=1}^{F}
\frac{
v_k^{x}\cdot \tilde{v}_k^{x} \;+\; v_k^{y}\cdot \tilde{v}_k^{y}
}{
\sqrt{(v_k^{x})^{2}+(v_k^{y})^{2}}
\;\cdot\;
\sqrt{(\tilde{v}_k^{x})^{2}+(\tilde{v}_k^{y})^{2}}
},
\end{equation}

\begin{figure*}[t]
  \centering
  \includegraphics[width=1.0\textwidth]{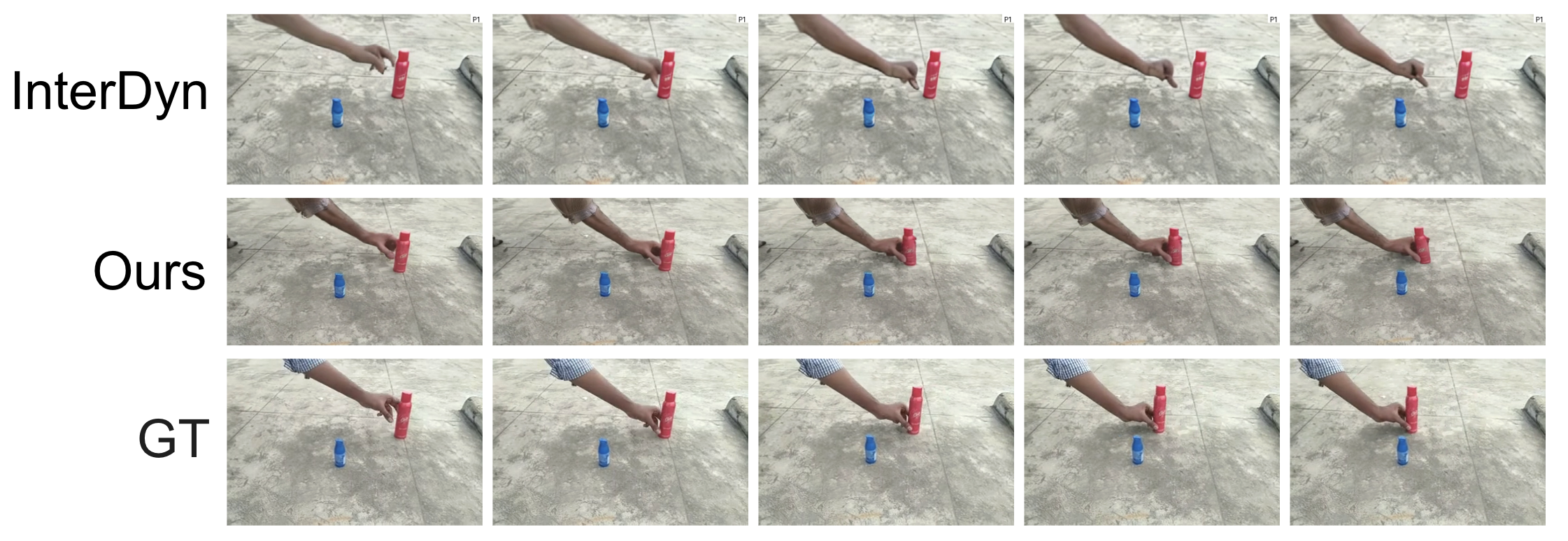}
  \caption{Qualitative Comparisons}
  \label{fig:QC_1}
\end{figure*}

\begin{figure}[t]
  \centering
  \includegraphics[width=\columnwidth]{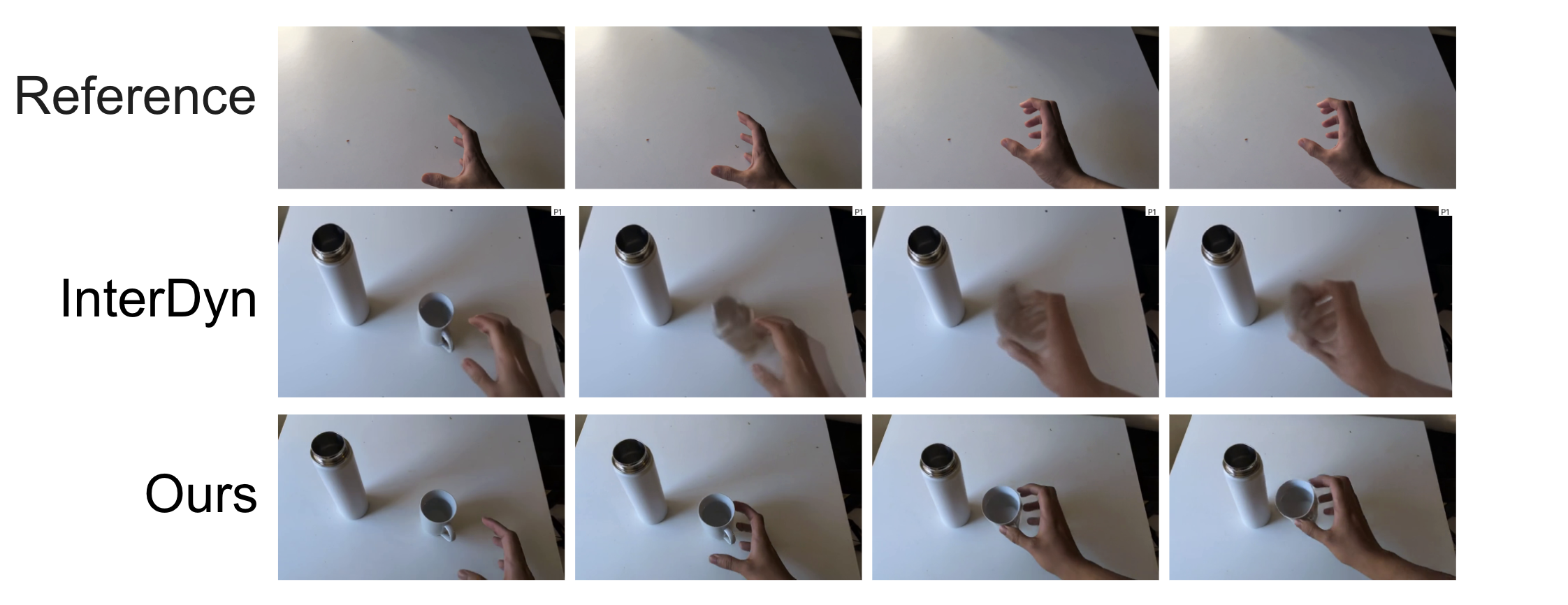} 
  \caption{Qualitative Comparisons with Reference}
  \label{fig:QC_2}
\end{figure}

\subsection{Baseline}
We compare with InterDyn as the alternative hand pose conditioned interaction simulation model and demonstrate our model significantly outperforms in pixel similarity, motion fidelity, and hand projection accuracy. InterDyn employs a controlnet to condition a video diffusion model on 2D masks to learn force propagation and dynamics. While employing a 2D masks allow a more general approach for force dynamics simulation, it lacks explicit hand conditioning and spatial awareness of the environment. 

\begin{table}[t]
\centering
\small
\setlength{\tabcolsep}{4pt}
\renewcommand{\arraystretch}{1.15}

\begin{tabularx}{\linewidth}{>{\raggedright\arraybackslash}X *{4}{>{\centering\arraybackslash}X}}
\toprule
Method &
\makecell{SSIM\\$\uparrow$} &
\makecell{PSNR\\$\uparrow$} &
\makecell{LPIPS\\$\downarrow$} &
\makecell{PCK@20\\$\uparrow$} \\
\midrule
DexSIM & 0.526 & 12.21 & 0.362 & 65 \\
w/o Spatial Cache & 0.506 & 11.58 & 0.374 & 65 \\
\bottomrule
\end{tabularx}
\caption{Spatial cache ablation comparison.}
\label{tab:ablation_latency_quality}
\end{table}

\subsection{Quantitative Comparisons}
DexSIM outperforms InterDyn on visual fidelity, motion smoothness and hand projection accuracy. PCK@20 determines the percentage of keypoints that are within 20 pixels of the ground truth. With spatial cache and depth signals, the object maintains better consistency throughout the generated frames as can be seen in Fig.~\ref{fig:QC_1}. And the explicitly conditioned hand pose allow demonstrate much improved hand projection accuracy compared to the baseline. 

Notably there is a quality and latency trade-off between the bi-directional model and causal versions of DexSIM. While the bi-directional model learns directly from the ground truth, the causal model's performance can be attributed to two factors: 1) multi-stage training runs and 2) inference time error accumulation. The causal model underperforms the bi-directional model on frame similarity, motion and hand projection metrics as the causal distillation training process relies on the bi-directional model as the teacher model and during inference, error is accumulated overtime as a frame-wise autoregressive model where as the bi-directional model generates all frames at once. However, the causal model is able to generate with 4 NFEs with an order of magnitude faster FPS in 15.24 compared to bi-directional model's 50 NFEs and 1.44 FPS along with much faster end-to-end time to first frame.

\subsection{Qualitative Comparisons}
Fig.~\ref{fig:QC_1} demonstrates the qualitative results of DexSIM and it's ability to conduct specially accurate dexterous simulations compared to InterDyn. DexSIM is able to accurately target the object and generate realistic simulations and interactions while following the hand action guidance signal. 2D signals such as segment masks have innate ambiguity as conditions passed to the diffusion models therefore, InterDyn fails at grasping objects with 2D mask conditioning more frequently. 

Fig.~\ref{fig:QC_2} we show motion transfer as an application of DexSIM where a motion from a source can be used for a different target video while manipulating and interacting with the object. Reference image and motion transfer is much more challenging with depth ambiguity of the environment and the hand signal leading to blurry generations by the baseline.


\begin{table}[t]
\centering
\small
\renewcommand{\arraystretch}{1.15}
\setlength{\tabcolsep}{10pt}

\begin{minipage}{0.72\linewidth} 
\centering
\begin{tabular}{@{}lccc@{}}
\toprule
\textbf{Methods} & \textbf{NFE} & \textbf{FPS} $\uparrow$ & \textbf{TTFF} $\downarrow$ \\
\midrule
DexSIM Bi-directional & 50 & 1.44  & 57.24 \\
DexSIM Causal         & 4  & 15.24 & 2.12  \\
\bottomrule
\end{tabular}
\end{minipage}

\caption{Runtime and latency comparison. TFF indicates
the end-to-end time-to-first-frame. }
\label{tab:runtime_latency}
\end{table}

\subsection{Ablation}
\textbf{Spatial cache} We conduct ablation experiments to investigate the significance of spatial cache for spatial memory and long-term consistent dexterous interaction generation. Spatial cache allows more accurate manipulation with objects with additional depth conditioning resulting in better performances for pixel and perceptual similarity where as we found the hand projection accuracy is similar due to the hand projection in 2D space not requiring spatial information.

\textbf{Causal Distillation} Distilling with DMD with bi-directional DexSIM as teacher to generate frames autoregressively results in a low-latency causal DexSIM with greater throughput with less number of function evaluations (NFE) resulting in a distilled 4-step model for real-time interactive dexterous manipulation simulation. And the FPS is significantly increased with end-to-end time-to-first-frame is greatly reduced through causal distillation. 

\section{Conclusion}
Here we present DexSIM a real-time interactive dexterous manipulation simulation model train from a 2-stage training process with a unified feature representation and roll-out training with spatial cache to provide spatial conditions for more accurate hand-object interaction.
While DexSIM focuses on improvements in spatial aware and memory generation, the scale of dataset and model size limits the model's performance in overall realism and appearance. Scaling compute with stronger base model and model and dataset size will significantly increase appearance realism and motion smoothness. 
Joint video and step distillation could be employed for reducing latency and denoising steps. And data free distillation to learn the teacher distribution better without overfitting on the training-data. Finally the spatial cache could be updated with more intricate techniques such as optical flow to preserve the depth information more semantically accurately. 
{
    \small
    \bibliography{iclr2026_conference}
    \bibliographystyle{iclr2026_conference}
}


\end{document}